\documentclass[9pt]{article}
\usepackage{listings}
\usepackage{amsthm}
\usepackage{amsmath}
\usepackage{amsfonts}

\usepackage{url}
\usepackage{graphicx}

\usepackage{listings}
\DeclareMathOperator*{\supp}{supp}

\title{Adversarial network training using higher-order moments in a modified Wasserstein distance}
\author{Oliver Serang\\
  A-Alpha Bio, Seattle, WA, USA\\
  oserang@aalphabio.com}

\begin{document}

\maketitle

\begin{abstract}
\noindent Generative-adversarial networks (GANs) have been used to
produce data closely resembling example data in a compressed, latent
space that is close to sufficient for reconstruction in the original
vector space. The Wasserstein metric has been used as an alternative
to binary cross-entropy, producing more numerically stable GANs with
greater mode covering behavior. Here, a generalization of the
Wasserstein distance, using higher-order moments than the mean, is
derived. Training a GAN with this higher-order Wasserstein metric is
demonstrated to exhibit superior performance, even when adjusted for
slightly higher computational cost. This is illustrated generating
synthetic antibody sequences.
\end{abstract}

\flushbottom
\maketitle
\thispagestyle{empty}

\section{Introduction}
\subsection{Generative-adversarial network}
The generative-adversarial network (GAN) is a game-theoretic technique
for generating values according to a latent distribution estimated on
$n$ example data $x \in \mathbb{R}^{n\times \ell\times
  u}$.\cite{goodfellow2014generative} GANs employ a generator, $g :
\mathbb{R}^y \rightarrow \mathbb{R}^{\ell\times u}$, which maps
high-entropy inputs to an immitation datum; these high-entropy inputs
$\in \mathbb{R}^y$ effectively determine a location in the latent
space and are decoded to produces an immitation datum. GANs also
employ a discriminator, $d : \mathbb{R}^{\ell\times u} \rightarrow
[0,1]$, which is used to evaluate the plausibility that a datum is
genuine. Generator and discriminator are trained in an adversarial
manner, with the goal of reaching an equilibrium where both implicitly
encode the distribution of real data in the latent space. If training
is successful, $\hat{X} = g\left(Z\right)$ (where $Z\sim
\mathcal{N}(0,1)^y$) will produce data resembling a row of $x$;
$d\left( \mathbb{R}^k \right)$ will correspond to the cumulative
density in a unimodal latent space where the latent space density
projects the empirical distribution of $x_i$.

\subsection{Cross-entropy loss}
GANs are typically trained using a cross-entropy loss to optimize the
parameters of both $g$ and $d$, which measures the expected bits of
surprise that samples from a foreground distribution would produce if
they had been drawn from a background distribution. The parameters
$\theta_g$ are optimized to minimize the surprise of the Bernoulli
distribution $1-\Sigma, d(g(Z))$ given the background distribution $0,
1$ (\emph{i.e.}, minimizing the surprise from a background that scores
$d(\hat{X})=1$). The parameters $\theta_d$ are optimized to minimize
the surprise of the Bernoulli distribution $1-\Sigma, d(x)$ given the
background distribution with $d(x_i)=1$ and $d\left(\hat{X}\right)=0$.

% WGAN
\subsection{Wasserstein metric loss}
When two distributions are highly dissimilar from one another, their
support may be distinct such that cross-entropy becomes numerically
unstable. This causes uninformative loss metrics: two distributions
with non-overlapping support are quantified identically to two
distributions whose supports are non-overlapping and very far from one
another. These factors lead to poor training, particularly given that
$g$ will initially produce noise, which will quite likely have poor
overlap with real data in the latent space.

For this reason, Wasserstein distance was proposed to replace
cross-entropy.\cite{arjovsky2017wasserstein} Wasserstein distance is
the continuous version of the discrete earth-mover distance, which
solves an optimal transport problem measuring the minimal movements in
Euclidean distance that could be used to transform one probability
density to another. Earth-mover distance is well defined, even when
the two distributions have disjoint support. This avoids modal
collapse while training.

If earth-mover distance is used to measure the distance between
distributions $p_A$ and $p_B$, then the set of candidate solutions
$\gamma$ will be functions with domain $\supp(p_A)\times \supp(p_B)$
and where the marginals equal $p_A$ and $p_B$. Thus, $\Delta_{EM}(p_A,
p_B) = \inf_{\gamma\in\Pi(p_A,p_B)} \mathbb{E}_{a,b \sim \gamma} \|a -
b\|$, where $\Pi(p_A,p_B)$ is the set of distributions with marginals
$p_A, p_B$.

The discrete formulation can be solved combinatorically via LP;
however, the continuous formulation, Wasserstein distance, is computed
via the Kantorovich-Rubinstein dual\cite{thickstun2019kantorovich},
which we show below.
\begin{eqnarray*}
  \Delta_W(p_A, p_B) &=& \inf_{\gamma\in \Pi(p_A,p_B)} \mathbb{E}_{a,b\sim \gamma} \|a - b\|\\
  &=& \inf_\gamma \mathbb{E}_{a,b\sim \gamma} \|a - b\| + \begin{cases}
    0, & \gamma \in \Pi(p_A,p_B)\\
    \infty, & \text{else}.
  \end{cases}
\end{eqnarray*}

The penalty term, here named $\lambda(p_A,p_B,\gamma)$, can be recreated using an adversarial critic function, $f$,
which has a unitless codomain:
\begin{multline*}
  \lambda(p_A,p_B,\gamma) =\\
  \sup_f \mathbb{E}_{a'\sim p_A} \left[f(a')\right] - \mathbb{E}_{b'\sim p_B} \left[f(b')\right] - \mathbb{E}_{a,b\sim \gamma} \left[f(a)-f(b)\right] =\\
  \begin{cases}
    0, & \gamma \in \Pi(p_A,p_B)\\
    \infty, & \text{else.}
  \end{cases}
\end{multline*}
$\lambda(p_A,p_B,\gamma)=\infty$ is achieved when $\gamma\not\in\Pi(p_A,p_B)$
because $f$ can be made s.t., w.l.o.g., $|f(a)|\gg 1$ at the value $a$
where $p_A(a) \neq \int_\infty^\infty \gamma(a,b)\partial b$.

Thus,
\begin{multline*}
  \Delta_W(p_A, p_B) =\\
  \inf_\gamma \sup_f \mathbb{E}_{a,b\sim \gamma} \left[ \|a - b\| + f(b) - f(a) \right] + \mathbb{E}_{a'\sim p_A}\left[f(a')\right] - \mathbb{E}_{b'\sim p_B}\left[f(b')\right].
\end{multline*}

We can further reorder $\inf_\gamma \sup_f$ to $\sup_f \inf_\gamma$:
For any function $t$, $h(\beta) = \inf_a t(\alpha,\beta)$, and $\delta
= \inf_\alpha \sup_\beta t(\alpha,\beta) = \inf_\alpha h(\beta)$, and
so $\forall \alpha, h(\beta) \leq t(\alpha,\beta)$. Thus $\inf_\alpha
\sup_\beta t(\alpha,\beta) \geq \inf_\alpha \sup_\beta h(\alpha) =
\inf_\alpha \delta = \delta$ (\emph{i.e.}, weak duality). Furthermore,
if $t$ is convex in $\alpha$ and concave in $\beta$, then the minimax
principle yields $\inf_\alpha \sup_\beta t(\alpha,\beta) = \sup_\beta
\inf_\alpha t(\alpha,\beta)$ (\emph{i.e.}, strong duality). Because
$\Delta_W$ is convex in $\gamma$ (here manifest via convexity in
$a,b$) and concave in $f$ (manifest via concave uses of $f$ rather
then concavity of $f$ itself), we have
\begin{multline*}
\Delta_W(p_A, p_B) =\\
\sup_f \inf_\gamma \mathbb{E}_{a,b\sim \gamma} \left[ \|a - b\| + f(b) - f(a) \right] + \mathbb{E}_{a'\sim p_A}\left[f(a')\right] - \mathbb{E}_{b'\sim p_B}\left[f(b')\right].\\
= \sup_f \mathbb{E}_{a'\sim p_A}\left[f(a')\right] - \mathbb{E}_{b'\sim p_B}\left[f(b')\right] + \inf_\gamma \mathbb{E}_{a,b\sim \gamma} \left[ \|a - b\| + f(b) - f(a) \right].
\end{multline*}

$\inf_\gamma$ is achieved by concentrating the mass of $\gamma$ where
$\|a - b\| + f(b) - f(a) < 0$ and setting $\gamma=0$ wherever $\|a -
b\| + f(b) - f(a) \geq 0$. Thus $\inf_\gamma \mathbb{E}_{a,b\sim
  \gamma} \left[ \|a - b\| + f(b) - f(a) \right] \leq 0$. This
constrains that where $\frac{f(a) - f(b)}{\|a-b\|}>1$, the dual
penalty term will become $-\infty$, and so we need only consider $f$
s.t. $\frac{f(a) - f(b)}{\|a-b\|}\leq 1$. This is equivalent to
constraining $f$ s.t. all secants having a maximum slope $\leq 1$
(\emph{i.e.}, Lipschitz $\|f\|_L\leq 1$) yields the weakest penalty,
0:
\[
\Delta_W(p_A, p_B) = \sup_{f:\|f\|_L\leq 1} \mathbb{E}_{a'\sim p_A}\left[f(a')\right] - \mathbb{E}_{b'\sim p_B}\left[f(b')\right].
\]

In WGAN training, our critic functions as $f$, exploiting differences
between real and generated sequences. The critic loss function is
simply the difference between mean critic values of generated
sequences minus mean critic values of real sequences; minimizing this
loss will maximize discrimination, with real sequences awarded higher
critic scores. With the goal of attaining Lipschitz continuity on $f$,
we constrain its parameters $\theta_f$, clipping them to small values
$\in [-\tau, \tau]$ at the end of each batch step. A small enough
$\tau$ will ensure Lipschitz continuity for any finite
network. Furthermore, $\|f\|_L\leq 1 \leftrightarrow
\frac{1}{\zeta}\|f\|_L\leq \zeta$; therefore, $\tau$ can be chosen
rather arbitrarily as long as $\tau\ll 1$, because the cone of
functional solutions $\{f : \|f\|_L\leq \zeta \leq 1\}$ includes all
nonnegative scales of functions for which $\|f\|_L\leq 1$. Choice of
$\tau$ will influence the optimal choice of learning rate.

When training $g$, the WGAN attempts to fool the critic and thus
maximize the loss used by the critic $f$. Thus, $g$'s loss is the
negative of $f$'s loss. In practice, $\theta_g$ does not influence
critic values of real data $f(x_i)$, and so $g$'s loss needs only be
$-\mathbb{E}\left[f(\hat{X})\right]$ to maximize critic scores of
generated sequences. 

All expectations are taken via Monte Carlo (\emph{i.e.}, by taking the
mean of $f$ scores over each batch).

\section{Methods}
\subsection{A WGAN using Wasserstein distance with higher moments}
In this manuscript, we propose a modified WGAN, in which we consider
other $\lambda'$ satisfying
\[
\lambda'(p_A,p_B,\gamma) = \begin{cases}
  0, & \gamma \in \Pi(p_A,p_B)\\
  \infty, & \text{else.}
\end{cases}
\]

Wasserstein distance employs duality via an adversarial $f$ that
concentrates where (w.l.o.g.) $p_A(a) \neq \int_\infty^\infty
\gamma(a,b)\partial b$:
\[
\lambda(p_A,p_B,\gamma)=\sup_f \mathbb{E}_{a'\sim p_A}
\left[f(a')\right] - \mathbb{E}_{b'\sim p_B}\left[f(b')\right] -
\mathbb{E}_{a,b\sim \gamma}\left[f(a)-f(b)\right].
\]
This correspond to $f$ exploiting deviations in the first moment of
$f$ under distribution $p_A$ (w.l.o.g.).

Motivated by the method of moments, we consider the first $m$
moments, $\mu_1, \mu_2, \ldots \mu_m$. At WGAN convergence, deviations
between $p_A,p_B$ and marginals of $\gamma$ should not be exploitable
at any $q$ moment:
\[
\lambda'_q(p_A,p_B,\gamma)=\sup_{f} \mathbb{E}_{a'\sim p_A}
\left[{f(a')}^q\right] - \mathbb{E}_{b'\sim p_B}\left[{f(b')}^q\right] -
\mathbb{E}_{a,b\sim \gamma}\left[{f(a)}^q-{f(b)}^q\right]=0.
\]
We continue using a signed deviation for the first moment
(\emph{i.e.}, $f(\hat{X})\leq f(x_i)$), but unsigned deviations for
the remaining moments:
\[
\lambda'(p_A,p_B,\gamma)=\lambda'_1(p_A, p_B,\gamma) + \sum_{j=2}^m
|\lambda'_j(p_A, p_B,\gamma)|.
\]
Note that $f$ is still used in a convex manner, as its outputs are
either unconstrained (in $\lambda_1$) or within a bounded polytope
$\lambda'_{j>1}$; strong duality holds.

The same derivation holds under central moments, which are used in
these experiments.

Lipschitz continuity under higher moments of $f$ is achieved by
decreasing $\tau$ s.t. $|f|\leq\epsilon\ll \frac{1}{2}$.  In this case, both
$\mu_1$ and the all central moments can be bounded: $\forall
j>1,|f^j-\mu_j|\leq 2\epsilon\ll 1$, and thus $\forall j>1,
\frac{{f(a)}^j - {f(b)}^j}{\|a - b\|} < \frac{f(a) - f(b)}{\|a - b\|}
< 1$. Thus Lipschitz continuity is ensured by the standard Wasserstein
derivation.

Since $f$ concentrates at deviations between distributions, it should
approach a Dirac delta before convergence if $g$'s training lags the
training of $f$; thus, here we have not investigated using the higher
moments informing $\lambda'$ when training $f$; $f$ is trained using
the same Wasserstein loss using $\lambda$. $\lambda'$ is used to train
$g$, which corresponds to replacing standard code {
  \footnotesize
\begin{lstlisting}
  preds_neg = critic(ins_neg)
  mean_neg = torch.mean(preds_neg)
  loss_gen = -mean_neg
\end{lstlisting}
}
with new code
{ \footnotesize
\begin{lstlisting}
  preds_neg = critic(ins_neg)
  preds_pos = critic(ins_pos)
  mean_pos = torch.mean(preds_pos)
  mean_neg = torch.mean(preds_neg)

  # signed difference between mu_1 values:
  lambda_gen = mean_pos - mean_neg
  for j in range(2, n_moments+1):
    loss_gen += torch.abs(
                # absolute difference between mu_j values:
                torch.mean( (preds_neg-mean_neg)**j ) -
                torch.mean( (preds_pos-mean_pos)**j )
                )
\end{lstlisting}
}

This formulation allows learning from batches in gestalt. When the
number of computed moments $m$ equals the batch size $b$, there is
sufficient information to recover the entire distribution;
furthermore, even relatively few moments can accurately summarize the
distribution in practice in a manner reminiscent of the fast multipole
method\cite{greengard1987fast, pfeuffer2016bounded}.

\subsection{Impact on runtime}
Considering higher moments at batch size $b$ results in a per-batch
runtime $\in \Omega(b^2)$ or $\in \Omega(m\cdot b)$ if the layer of
moments is used to separate the data from critic output. Also, the
modified WGAN needs to compute critic scores $f(x_i)$ when training
the generator. This increases computation cost. 

Fractional moments can be informative and numerically stable; however,
in the general case, they require arithmetic on complex numbers and
may negatively influence performance.

\section{Results}
% WGAN to create a synthetic antibody library
To benchmark WGAN training methodology, we train WGANs to output heavy
chain antibody sequences. The overall scheme for this WGAN is heavily
inspired by the seminal neural network-based antibody sequence design
work of Tileli \emph{et al.}\cite{tileli2020designing}; furthermore,
our model architecture is inspired by the multi-layer convolutional
network from that work.

\subsection{Experimental setup}
\subsubsection{Sequence data}
Heavy chain sequence examples from from Observed Antibody
Space\cite{kovaltsuk2018observed, olsen2022observed} are filtered for
outliers based on sequence length and sampled to $2.5\times 10^5$
sequences. Note that sequences are not embedded using a multiple
sequence alignment; instead, every sequence is appended with starting
and ending characters \^{} and \$ and then padded with \$ so that all
sequences have the same length. Sequences are embedded via one-hot
embedding. 

\subsubsection{Critic and generator model architectures}
{\bf Critic:} The critic is constructed of two 2D convolutional
layers. For simplicity, 2D convolution is performed with padding such
that there is no movement of the kernel over the axis labeling amino
acids in the one-hot embedding and the input padded sequence length
$\ell$ equals the length of the convolved vector. In this manner, each
of the $c$ channels of the first 2D convolution essentially passes a
PSSM with $u$ embedding characters over the sequence. These $c$
channels are transposed to view them as a single matrix of one channel
with an alternate embedding with $c$ characters. This is again
nonlinearized with leaky ReLU, and 2D convolved again to produce a
single channel output. This is now akin to using a PSSM on $k$-mer
motifs (using an alphabet of $c$ possible motifs) rather than on amino
acids, which is in turn equivalent to inferring an order-$k-1$ Markov
model. Padding is performed in the same manner as in the first 2D
convolution; the output is a vector of the same length as the original
amino acid sequence. This vector is condensed to a single value via
feedforward layers: each collection has linear layers that halve the
number of nodes followed by leaky ReLU transfer function to allow
nonlinearity. Note that 2D convolution here is equivalent to several
channels of 1D convolution and can be implemented as such.

{\bf Generator:} The generator is nearly identical to the critic in
reverse. Thinking of the critic and generator as two halves of an
autoencoder, inverting the critic's compression to lower-dimensional
latent space, deconvolution would be desired; however, deconvolution
is a form of convolution (but with a kernel whose values have been
multaplicatively inverted in the frequency domain) as shown by the
convolution theorem.\cite{proakis2001digital}

A standard normal noise vector $Z$ inflates to a vector with length
$\ell\cdot u$ where $u$ is the size of the alphabet used for one-hot
embedding. A leaky ReLU is used to permit nonlinearity. The vector is
then viewed as a matrix $\in\mathbb{R}^{\ell\times c}$ matrix and is
convolved in 2D to produce $c$ channels of output (padding in the 2D
convolution matches the approach used in the critic). This output is
transposed to be viewed as a single matrix of one channel with an
alternate embedding in $c$ new characters. Nonlinearity is again induced
with leaky ReLU. The matrix is then convolved 2D again with the same
padding strategy and $u$ channels out and transposed to form a matrix
of one channel and $u$ characters embedded. This is nonlinearized with
leaky ReLU. Note that this matrix is of the same shape as used by the
sequence embedding. Softmax is then applied to the character embedding
axis, forcing it into an embedding that resembles a one-hot.

All leaky ReLUs have negative slope 0.2. Layers are delimited with
dropout 0.1 during training, but not during evaluation.

\subsubsection{Evaluation}
After each epoch, quality of $\hat{x}$ are evaluated using KL
divergence of the categorical distributions of 6-mer sequences given
the 6-mer sequence distribution from the $2.5\times 10^5$ heavy chain
antibody sequences and $2\times 10^4$ sequences sampled from $g$. For
numeric stability, KL divergence is computed using a pseudocount of
$10^{-10}$ added to values not in the background distribution's
support.

\subsubsection{Hyperparameters}
Learning rate and batch size are chosen by training a standard WGAN
network is trained for several replicates with various learning rates
$\in \{0.1, 0.01, 0.001, 0.0001\}$ and batch sizes $b \in \{64, 128,
256, 512\}$. The learning rate that produced the best 6-mer KL
divergence is 0.001. $b=128$ yielded the best 6-mer KL divergence
while still maintaining $>75\%$ GPU usage with {\tt nvidia-smi}. These
hyperparameters are used for training the WGANs using higher-order
moments.

\subsubsection{Reproducibility}
The random seeds $0, 1, \ldots 4$ are used for replicate
experiments. This includes seeding {\tt random}, {\tt numpy.random},
{\tt torch}, and {\tt torch.utils.data.DataLoader}. {\tt
  torch.use\_deterministic\_algorithms(True)} is used, along with the
accompanying recommended environment variable set by {\tt export
  CUBLAS\_WORKSPACE\_CONFIG=:4096:8}.

\subsubsection{Training details}
Models are instantiated and trained with pytorch 1.10. A shuffled {\tt
  DataLoader} with 8 worker threads and pinned memory is used in
training. In each epoch, $d$ is trained on each batch and $g$ is
trained on $\frac{1}{5}$ of batches to avoid adjusting $g_\theta$ with
improper guidance from an uninformed critic.

Adaptive moment estimation (Adam) is used for gradient descent.\cite{kingma2014adam}

Benchmarks are performed on AWS {\tt g4dn.8xlarge} instance using a
single Nvidia T4. Storage IOPS and throughput are maximized.

\subsection{Influence of higher moments on WGAN performance}
Data are produced using 5 replicate trials of 200 epochs. For
fairness, each loss function investigated used every random seed $\in
0, 1, \ldots 4$.

\begin{figure}
  \centering
  \includegraphics[width=5in]{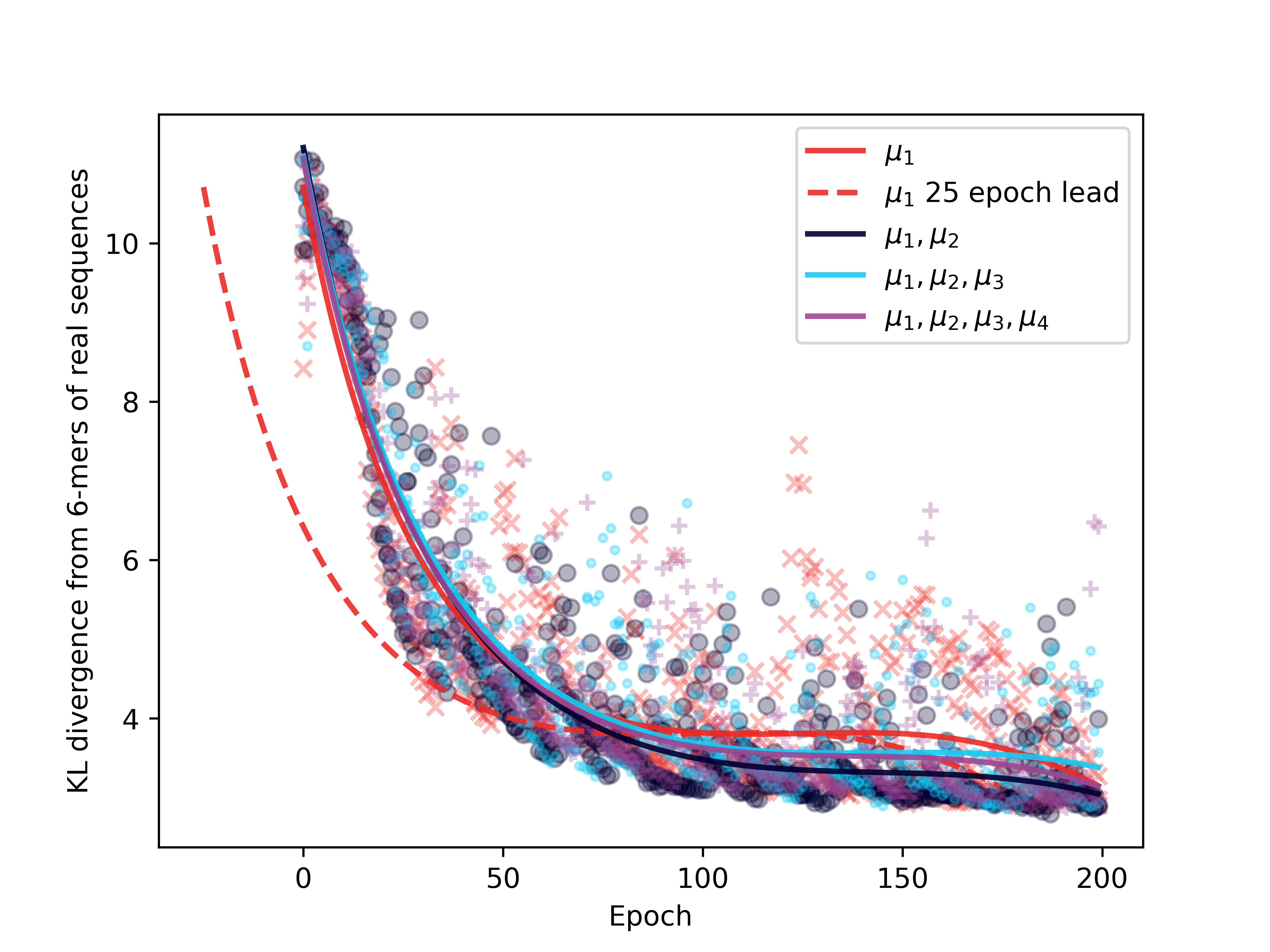}
  \caption{{\bf Sequence quality during training.} A standard WGAN
    (with loss using only the first moment $\mu_1$) is compared to
    WGANs using further central moments $m \leq 4$. Each scatter data
    point represents an estimate of the post-epoch 6-mer KL divergence
    on $2\times 10^4$ sequences generated by $g$. Five replicate
    experiments are performed for each model, each is fit with a
    curve, and the aggregate of all curves is plotted. For reference,
    the dashed line plots the standard WGAN shifted 25 epochs
    early.\label{fig:seq_qual}}
\end{figure}

Figure~\ref{fig:seq_qual} illustrates the relationship between
sequence quality produced by $g$ at each epoch and the loss function
used.

\begin{figure}
  \centering
  \begin{tabular}{cc}
    \includegraphics[width=2.3in]{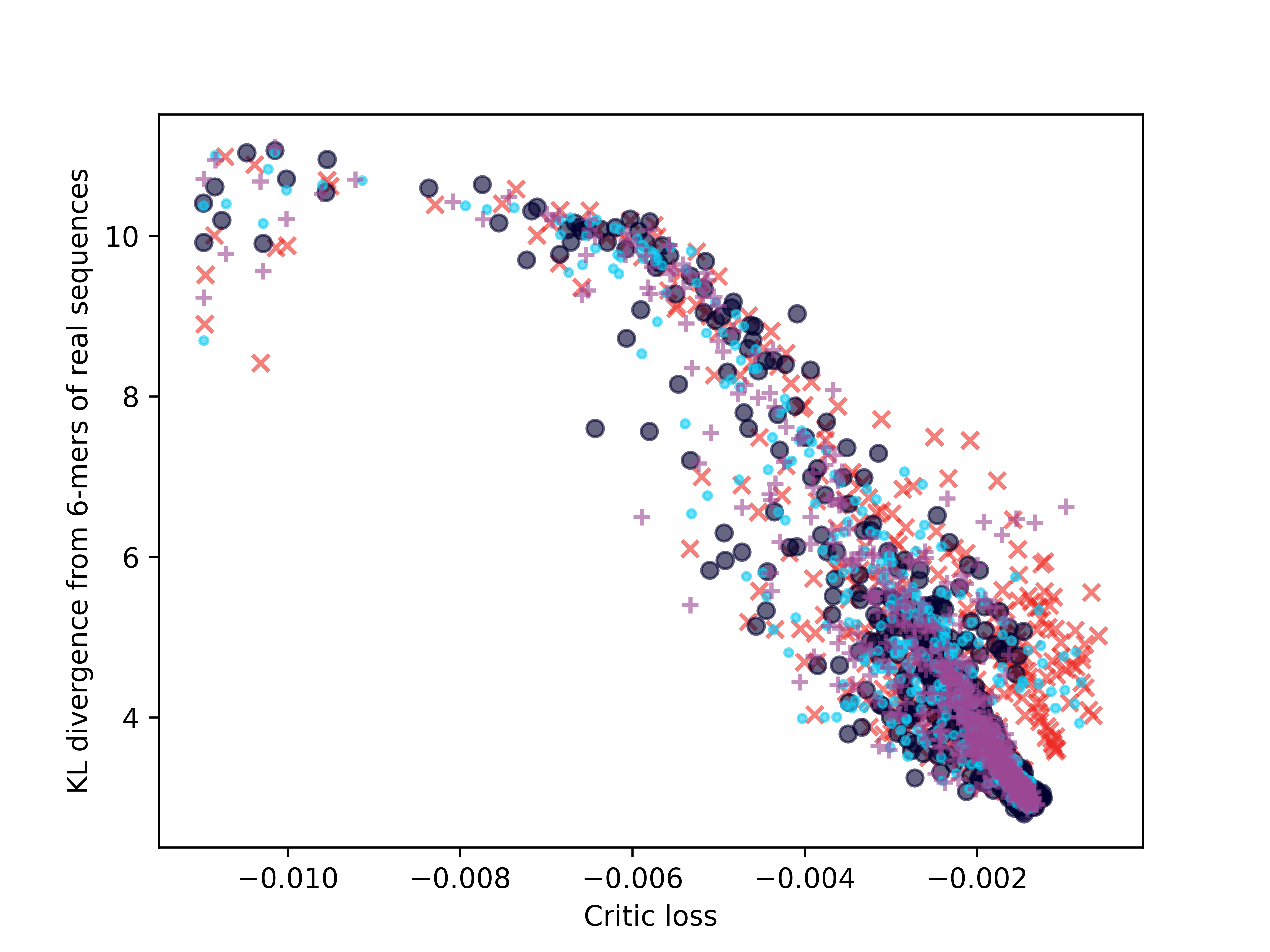} & \includegraphics[width=2.3in]{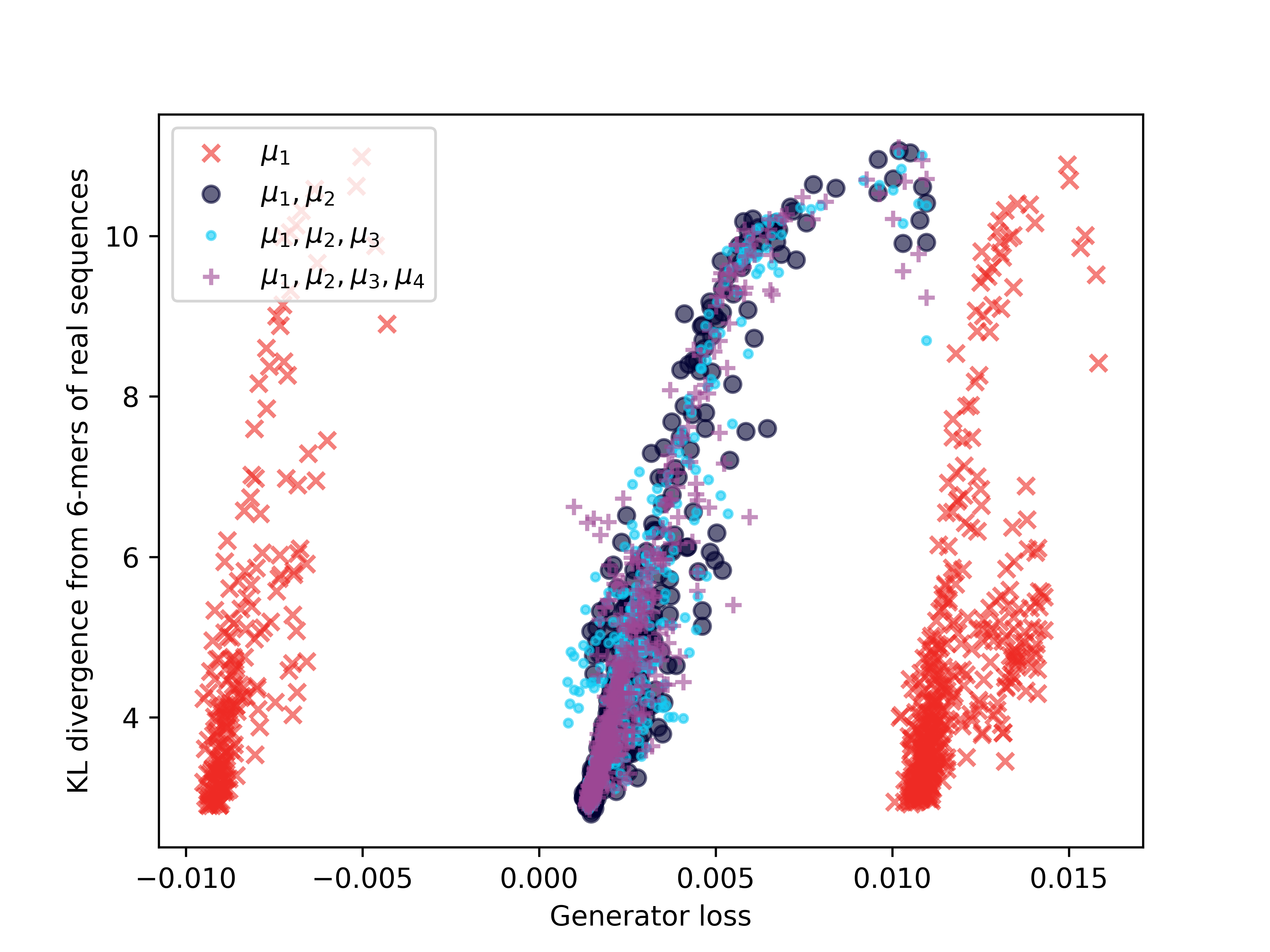}\\
    {\bf (a)} & {\bf (b)}
  \end{tabular}
  \caption{{\bf Quality of generator loss function throughout
      training.} A standard WGAN (with loss using only the first
    moment $\mu_1$) is compared to WGANs using further central moments
    $m \in \{1,2,3,4\}$ using data obtained from
    Figure~\ref{fig:seq_qual}. The correspondence between the loss
    functions and the 6-mer KL divergence quantify the quality of the
    loss function as a surrogate for optimizing sequence
    quality. Critic loss is standard throughout, but is affected by
    generator training; generator loss varies with $m$. Loss functions
    are computed during training, and so include dropout and other
    stochastic effects.\label{fig:loss_qual}}
\end{figure}

Figure~\ref{fig:loss_qual} illustrates the relationship between
sequence quality produced by $g$ and the loss function value for each
loss function used. Note that for any $m>1$, generator loss uses
penalty term $\lambda' \geq \lambda$, and so a small $m>1$ loss
function necessarily implies a small loss function using a standard
WGAN.

\begin{table}
  \centering
  \begin{tabular}{r|ccc}
    $m$ & Mean runtime (s) & KL \& crit. loss $\rho$ & KL \& gen. loss $\rho$\\
    \hline
    1 & 2917.66 & 0.7059 & 0.4169\\
    2 & 3079.00 & 0.8961 & 0.9257\\
    3 & 3097.12 & 0.8388 & 0.8722\\
    4 & 3110.64 & 0.8921 & 0.9205\\
  \end{tabular}
\caption{{\bf Runtime and quality of various generator loss
    functions.} Standard WGAN loss with $m=1$ is compared to losses
  with higher moments $m>1$. Total 200 epoch runtime is averaged over
  five replicates. The Spearman rank correlation coefficient $\rho$
  between each loss function and the 6-mer KL divergence is also
  displayed. Critic loss is standard throughout, but is affected by
  generator training. Generator loss varies $m\in \{1, 2, 3,
  4\}$. Loss functions are computed during training, and so include
  dropout and other stochastic effects.\label{tab:loss_qual}}
\end{table}

\section{Discussion}
% runtime:
Figure~\ref{fig:seq_qual} demonstrates that early in training, the
standard WGAN exhibits superior performance; however, later on, using
higher moments results in benefit to sequence quality, specifically
for the $m=2$ and $m=4$ loss functions. This is shown to be rougly
equivalent to gaining a 25 epoch advantage.

Figure~\ref{fig:loss_qual} and Table~\ref{tab:loss_qual} demonstrate a
greater correspondence between sequence quality and loss functions
with higher moments.

Qualitatively, using higher moments incentivizes optimizing batches as
a whole. One way that this may manifest is by improving batch
diversity of $\hat{x}$ to better match that of $x$, thereby reducing
modal collapse. For early epochs, this could explain the slightly
poorer performance, as these $m>1$ loss functions will initially be
less seeking of a dominant nearby mode.

Interestingly, $m=3$ did not perform as well as $m=2$ and $m=4$. This
could be because training the critic inherently drives $f$ toward a
high-concentration similar to a Dirac delta. While the first moment
$\mu_1$ is informative and even moments describe spread (variance
$\mu_2$ quantifies spread, excess kurtosis $\mu_4$ quantifies modality
near $\mu_1$), the skew, $\mu_3$, informs of direction but in a way
that may here be less useful or numerically stable than simply using
$\mu_1$.

Using higher moments increased runtimes, but not
substantially. Training modified WGAN with $m=2$ moments in $g$'s loss
required $<5.6\%$ more runtime than the standard WGAN. At a cost of
\$2.176 per hour\cite{aws_prices}, this corresponds to a cost of \$1.76
per replicate of the standard WGAN, and less than \$0.10 more
expensive to train the $m=2$ variant; however, the $m=2$ variant
reaches comparable convergence in 75\% of the training, and thus would
cost roughly \$1.63 per replicate. For larger data and more stringent
convergence criteria, the exponentially decaying gain in sequence
quality by training for further epochs suggests that this 25 epoch
advantage demonstrated by the $m>1$ variants would produce benefits
far more dramatic. Furthermore, it is likely the benefits illustrated
here would be stronger with more replicate experiments.

It is possible that deviations for different moments should receive
their own weighting in computing the loss function. In this manner, it
may be desirable to perform batch-based discrimination, where each
batch is reduced to its constituent moments, and then a critic
$f'(\mu_1, \mu_2, \ldots \mu_m)$ is computed on the moments $\mu_1,
\mu_2, \ldots \mu_m$ of critic values $f$ for the batch. Parameters
$\theta_{f'}$ could be learned and clamped during training to ensure
$\|f'\|_L\leq 1$.

\section{Conclusion}
Here we have shown that viewing distributions with several moments rather
than only using the first moment, $\mathbb{E}[f]$, improves WGAN
training. We could also easily train the critic $f$ using this strategy. 

\section{Acknowledgements}
Thank you to Ryan Emerson and Randolph Lopez for the scientific
discussion, James Harrang for the helpful comments, and to the entire
A-Alpha Bio team.

\section{Declarations}

\subsection{Conflicts of interest}
O.S. is an employee of A-Alpha Bio and owns stock options in the
company.

%\bibliography{refs}
%\bibliographystyle{unsrt}

\end{document}